\documentclass{article}

\PassOptionsToPackage{numbers, compress}{natbib}

\usepackage[preprint]{neurips_2026}
 \usepackage{graphicx}
 \usepackage{amsmath}
 \usepackage{amssymb}
 \usepackage{wrapfig}
\usepackage{titletoc}
\usepackage{minitoc}
\usepackage{algorithm}
\usepackage{algpseudocode}
\usepackage{float}

\usepackage[utf8]{inputenc} %
\usepackage[T1]{fontenc}    %
\usepackage{hyperref}       %
\usepackage{url}            %
\usepackage{booktabs}       %
\usepackage{amsfonts}       %
\usepackage{nicefrac}       %
\usepackage{microtype}      %
\usepackage[table]{xcolor}         %

\title{AdaState: Self-Evolving Anchors for Streaming Video Generation}

\author{
  Yusuf Dalva \qquad Pinar Yanardag
  \\
  \texttt{\{ydalva, pinary\}@vt.edu}
  \\
  Virginia Tech
  \\
  \url{adastate.github.io}
}

\begin{document}

\maketitle

\begin{figure}[h!]
    \centering
    \includegraphics[width=\linewidth]{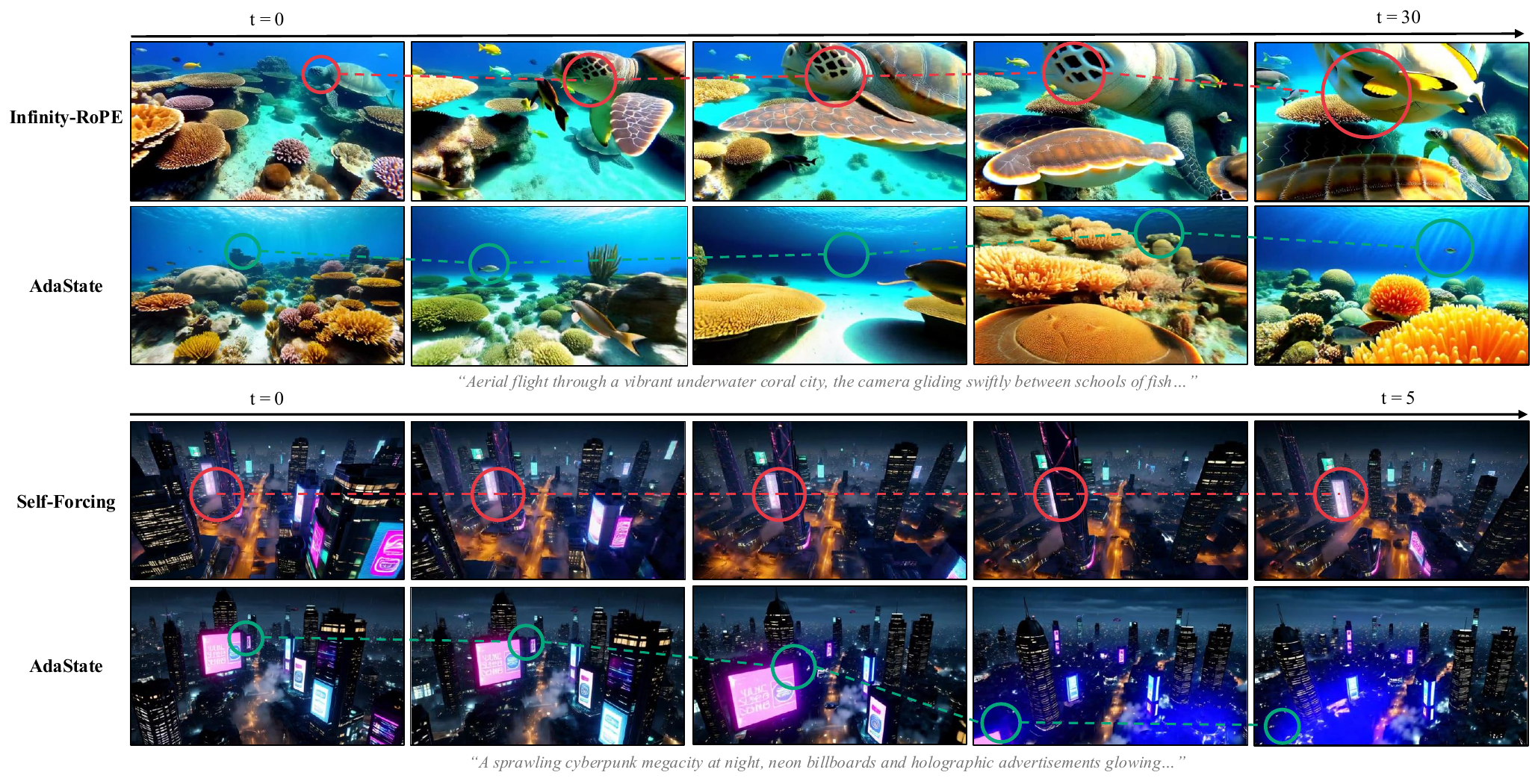}
    \caption{\textbf{AdaState.} Colored markers highlight the scene at each
timestamp; dashed lines trace their progression (\textcolor{red}{red}: baselines, \textcolor{teal}{teal}: AdaState). \textit{Top, t=30s (6$\times$ training horizon):} Infinity-RoPE's static anchor cannot adapt to the evolving scene, forcing the model to realize all implied content, schools of fish, sea turtles, within the initial layout, producing hallucinated duplications by t=30. AdaState's markers drift forward as new terrain and subjects emerge naturally across the rollout. \textit{Bottom, t=5s:} Self-Forcing holds the scene in place while \textit{neon lighting} dims and the \textit{cityscape} desaturates. AdaState produces continuous scene evolution while preserving visual characteristics.
}
    \label{fig:teaser}
\end{figure}

\begin{abstract}
Autoregressive video diffusion models generate streaming video by producing frames sequentially, conditioning each chunk on previously generated content. These models are structurally anchored to the first frame: its key-value representation occupies a privileged position in the attention cache and serves as the primary scene reference throughout generation. As the cleanest and most error-free position in the cache, this anchor draws disproportionate attention, suppressing video dynamics, and locking scene composition to the initial viewpoint even as the scene naturally evolves. The result is a temporally shallow video in which motion, camera movement, and scene progression are dampened in favor of static consistency. To address this, we replace the static anchor with an \emph{adaptive state}, a hidden latent that the model denoises alongside content at every chunk but never renders. Rather than referencing a frozen first frame, the model generates its own scene anchor at each step by attending to both the previous state and the current content, producing a reference that evolves with the generated content. Unlike standard video generation, which encodes an absolute notion of time, our formulation treats time as relative: every generation step sees the same positional structure regardless of how far generation has progressed, and the state transition is identical at every chunk. Together, these properties introduce a recurrence into the generation process, where denoising serves as the transition function, and the KV cache serves as the carrier, requiring no external module. Experiments demonstrate that the adaptive state substantially improves video dynamics, enabling richer motion and natural scene progression within generated videos.
\end{abstract}

\section{Introduction}
\label{sec:intro}
 
Autoregressive video diffusion models generate streaming video by producing one chunk of frames at a time, conditioning each chunk on previously generated content~\cite{huang2025selfforcing,yin2025causvid,yang2026longlive,liu2025rolling, yi2025deep}. These models are structurally anchored to the first frame: its key-value representation occupies a privileged position in the attention cache and serves as the primary scene reference throughout generation, exploiting the attention-sink phenomenon~\cite{xiao2024efficient} where causal softmax concentrates mass on initial positions. However, as the cleanest and most error-free position in the cache, this static anchor draws disproportionate attention, suppressing video dynamics and locking scene composition to the initial viewpoint even as the scene naturally evolves during generation. Motion, camera movement, and scene progression are dampened in favor of static consistency, producing temporally shallow video that lacks the dynamic richness of natural video (see Figure~\ref{fig:teaser}). The static anchor also limits the model's trained robustness, since errors accumulated over the autoregressive rollout are absorbed by the clean reference rather than surfaced at the model's most attended position, leaving the training objective with limited leverage to shape behavior under the imperfect conditions the model will encounter as generation extends.
 
Existing approaches do not address this root cause: static sinks~\cite{yang2026longlive,liu2025rolling} retain first-frame tokens as fixed anchors, reinforcing the shortcut. EMA-based methods~\cite{lu2025reward,kim2026memrope} apply content-agnostic averaging over evicted content, converging to a blurry mean that cannot adapt to scene changes. Token replacement approaches~\cite{li2026rolling} substitute raw cached content into the anchor position on a heuristic schedule, keeping the reference fresh but biasing generation toward reproducing past frames rather than producing novel continuation. All of these either preserve the static anchor or update it through operations external to the generative model, leaving the fundamental attention bias intact.
 
In this paper, we replace the static anchor with an \emph{adaptive state}, a hidden latent that the model denoises alongside content at every chunk but never renders. At each step, the model generates its own scene anchor by attending to both the current content and the previous adaptive state, producing a reference that evolves with the generated content rather than remaining locked to the initial frame. After denoising, the state's clean representation is written to the anchor position in the cache, while the content is decoded into video; the state is carried forward silently as an evolving scene reference. Unlike standard video generation, which encodes absolute temporal position, our formulation treats time as relative: every generation step sees the same positional structure regardless of how far generation has progressed, removing the notion of a privileged time zero. This design reveals that denoising is itself a recurrence: the adaptive state is a hidden variable updated by the model's own iterative refinement and carried via the KV cache, turning sequential autoregressive generation into a recurrent process with no external module or gating mechanism. To ensure the training objective emphasizes the frames that depend most on the adaptive state, we further propose horizon-weighted DMD, a per-frame loss weighting that increases with frame index, preventing the optimizer from concentrating capacity on clean, early frames. Experiments demonstrate that the adaptive state substantially improves video dynamics, enabling richer motion and natural scene progression within generated videos. \textbf{Our contributions:} (1) identifying the structural attention bias that suppresses dynamics in autoregressive video diffusion; (2) replacing the static anchor with an adaptive state trained via horizon-weighted DMD; and (3) showing that denoising is itself a recurrence, a hidden latent carried via the KV cache with no external module.

\begin{figure}
    \centering
    \includegraphics[width=\linewidth]{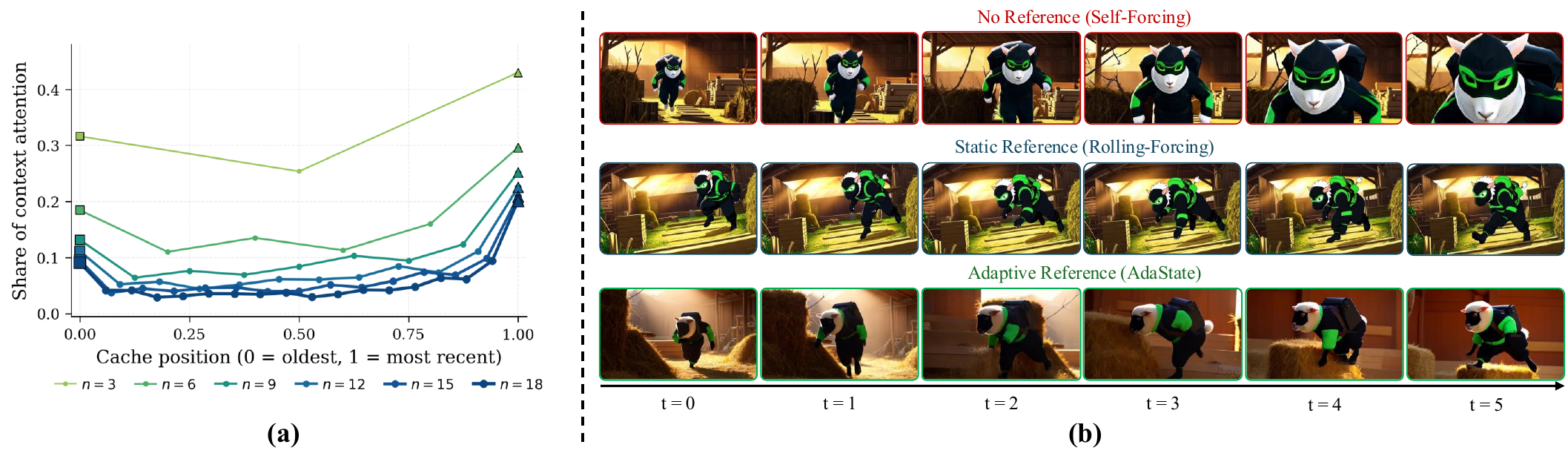}
    \caption{\textbf{The anchor-recency structure of streaming video attention.} (a) Off-diagonal attention in Self-Forcing across chunk depths. The anchor at position 0 (squares) and the freshest chunk frame (triangles) consistently dominate; remaining positions receive roughly uniform mass. (b) 5-second generation on the same prompt. Without a persistent reference, coherence degrades over time. A static reference preserves identity but freezes the scene. AdaState maintains identity while the camera moves and the environment evolves.}
    \label{fig:motivation}
\end{figure}

\section{Related Work}
\label{sec:related}
 
\noindent\textbf{Streaming autoregressive video diffusion.}
Autoregressive video diffusion models generate chunks sequentially with causal attention and KV caching for bounded-cost streaming. Distillation-based training has been the central thread: CausVid~\cite{yin2025causvid} introduced asymmetric distillation from bidirectional teachers, Self-Forcing~\cite{huang2025selfforcing} trains on model outputs via DMD to close the train-inference gap, Self-Forcing++~\cite{cui2026selfforcing} extends this to minute-scale, and Causal Forcing~\cite{zhu2026causal} refines the per-frame objective. Rolling Forcing~\cite{liu2025rolling} jointly denoises a rolling window, Reward Forcing~\cite{lu2025reward} adds reward-weighted distillation, and MMM~\cite{cai2026mode} couples flow matching on long videos with distribution matching on sliding windows. A common design retains first-frame KV as a static attention anchor, exploiting the attention-sink phenomenon~\cite{xiao2024efficient}; LongLive~\cite{yang2026longlive} makes this explicit by pinning the first frame's KV as a permanent sink. Approaches that update the anchor are content-agnostic: EMA over evicted tokens (Reward Forcing), dual-rate EMA with online RoPE re-indexing (MemRoPE~\cite{kim2026memrope}), raw-content replacement on a heuristic schedule (Rolling Sink~\cite{li2026rolling}), spatial-hierarchy compression (PackForcing~\cite{mao2026packforcing}), and block-relativistic positional encoding (Infinity-RoPE~\cite{yesiltepe2025infinity}). All either freeze the anchor or update it externally; AdaState updates it through the model's own denoising process, the same computation that produces content.

\noindent\textbf{Persistent state and test-time adaptation.}
RNNs, LSTMs, and modern selective state-space models~\cite{rumelhart1985learning, hochreiter1997long, chung2014empirical, gu2023mamba} maintain hidden states updated at each step by a learned transition function for bounded-cost sequence modeling. Test-Time Training (TTT)~\cite{sun2025learning} reframes recurrence by making the hidden state a model itself, updated via self-supervised learning at each step. Titans~\cite{titans} introduces a neural long-term memory module with adaptive forgetting; LaCT~\cite{lact} combines large-chunk TTT with sliding-window attention for video generation; VideoSSM~\cite{yu2025videossm} applies SSM-based global context to autoregressive video diffusion. Our work shares the structural property of a persistent hidden variable that is updated at each step and conditions the output without being observed directly. The distinguishing feature is the transition function: rather than a learned gate, recurrence matrix, SSM convolution, or gradient-based update, the state transition is the diffusion model's own multi-step denoising, a pretrained iterative refinement process repurposed as a recurrent update.

\noindent\textbf{Latent reasoning and thinking tokens.}
A growing line of work augments transformers with latent positions that carry intermediate computation, shaped by task loss rather than explicit supervision: Scratchpads~\cite{nye2022show} pioneered intermediate computation tokens for language models, Pause tokens~\cite{goyal2024think} showed that empty positions before output improve quality, and \cite{pfau2024lets} demonstrated filler tokens enable hidden computation. Coconut~\cite{hao2025training} feeds hidden states back as input embeddings in continuous space; CODI~\cite{shen2025codi} aligns latent states with token embeddings via distillation; Huginn~\cite{geiping2026scaling} uses a looped transformer with recurrent processing. Most recently, \cite{he2026reasoning} showed that latent tokens in diffusion language models, jointly predicted but never decoded, improve reasoning. Our adaptive state applies this principle to video generation: a latent slot processed alongside content, shaped by the generation loss, and serving as a queryable scene reference. The critical difference is persistence~--- language thinking tokens typically exist within a single forward pass, while our state persists across chunks via the KV cache and is updated through iterative denoising, functioning as a recurrent hidden variable that carries scene information across generation steps.

\begin{figure}
    \centering
    \includegraphics[width=\linewidth]{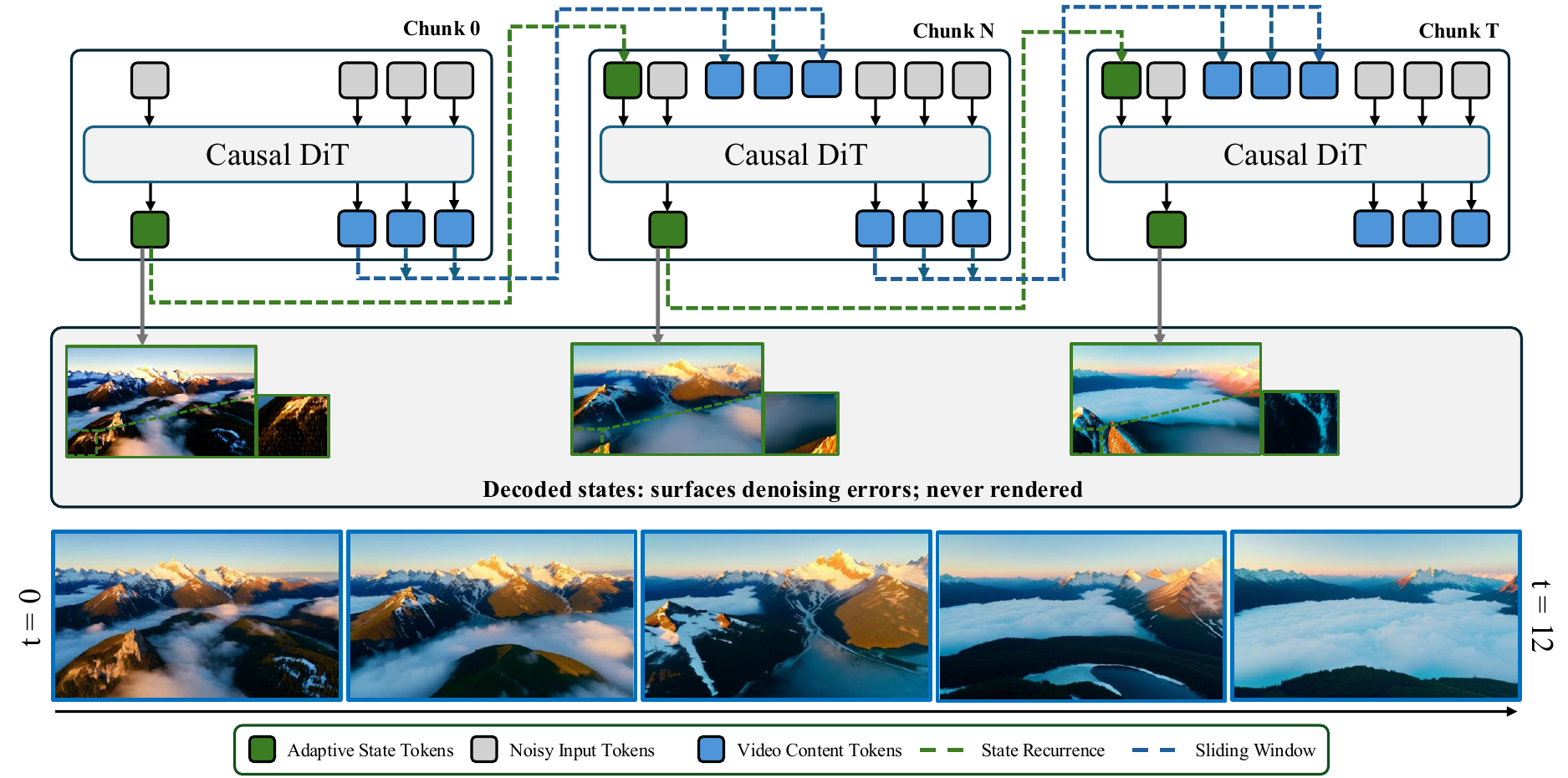}

    \caption{\textbf{AdaState Framework.} The adaptive state (green) is denoised alongside content at each chunk but never rendered. Its clean KV is written to position 0 and carried to subsequent chunks via the state recurrence (green dashed). Decoded state previews (middle, green-bordered, matching the state tokens) visualize the hidden state in image space; the zoom insets reveal the model's denoising errors, which the architecture surfaces at the cache anchor rather than letting a clean reference absorb them silently. The bottom strip (blue-bordered, matching the video content tokens) shows the cleanly denoised video output, with continuous scene progression from $t{=}0$ to $t{=}12$.}
    \label{fig:framework}
\end{figure}

\section{Method}
\label{sec:method}
 
\paragraph{Preliminaries.} We build AdaState on the autoregressive video diffusion framework of Self-Forcing~\cite{huang2025selfforcing}, where a student generator $G_\theta$ is distilled from a bidirectional teacher via Distribution Matching Distillation (DMD)~\cite{yin2024one}. Generation proceeds autoregressively in chunks of $F$ latent frames, each denoised from noise through a $K$-step schedule following the flow matching interpolation $\mathbf{x}_t = (1 - t)\mathbf{x}_0 + t\boldsymbol{\epsilon}$, $\boldsymbol{\epsilon} \sim \mathcal{N}(0, I)$~\cite{esser2024scaling}. At each chunk $n$, the generator denoises content $\mathbf{x}_c^{(n)}$ conditioned on a KV cache $\mathbf{h}_n$ of clean key-value pairs from prior chunks. The DMD loss drives this distillation by matching score functions between a frozen teacher $S_\text{real}$ and a learned critic $S_\text{fake}$ on the student's noised predictions, with gradient $\nabla_\text{DMD} = (S_\text{fake}(\hat{\mathbf{v}}_t, t) - S_\text{real}(\hat{\mathbf{v}}_t, t))/\gamma$ applied as a pseudo-target~\cite{yin2024one}. To enable generation beyond a fixed context length, the KV cache operates as a sliding window of size $W$, where keys are stored without positional encoding and re-encoded at read time with block-relativistic RoPE~\cite{yesiltepe2025infinity}, mapping the visible window to constant relative positions regardless of generation progress. As content exits the window, existing methods compensate by retaining the first frame's clean KV at a fixed sink position as a static scene anchor~\cite{yang2026longlive,liu2025rolling}, or by applying EMA over evicted tokens~\cite{lu2025reward}. Following Self-Forcing~\cite{huang2025selfforcing}, $G_\theta$ is trained by performing the full autoregressive rollout, so the student learns to generate from its own imperfect outputs.

\subsection{Context Utilization in Self-Forcing}
\label{sec:attention_analysis}
 
We probe Self-Forcing's attention over the cached KV window: for each denoising step, we measure post-softmax attention mass per cached K-frame, averaged over heads and renormalized over off-diagonal positions so all cached frames compete for the same mass budget. Figure~\ref{fig:motivation}(a) reveals a persistent bimodal structure across all chunk depths: the anchor at position~0 and the freshest chunk-summary frame consistently dominate, while the remaining ${\sim}70\%$ of cached positions receive roughly uniform attention. The anchor's absolute share decays with cache size, while its relative position ranks 2\textsuperscript{nd}-3\textsuperscript{rd} in all cases. Generation is thus driven primarily by what occupies these two positions, not by the full cached history. Figure~\ref{fig:motivation}(b) illustrates the consequence: in Self-Forcing, the first frame's KV remains at position~0 throughout generation, and its persistent attention mass constrains the scene from evolving naturally. Methods that pin a static anchor further amplify this, preserving identity but freezing the scene; an adaptive reference at the same position lifts the constraint, maintaining identity while the camera tracks and the environment evolves. Rather than redistributing attention across the cache, we intervene at the position the model already attends to most and replace its frozen content with an evolving, self-generated scene reference.
 
\subsection{Adaptive State}
\label{sec:state}
 
We replace the static first-frame anchor with an \emph{adaptive state} $s \in \mathbb{R}^{F_s \times H \times W \times C}$, a hidden latent that the generator denoises alongside content at every chunk but never renders (see Figure~\ref{fig:framework}). At each chunk $n$, the generator processes $F$ content frames (the current video chunk) together with $F_s$ state frames, forming the visible window. The content window spans $W_p{+}F$ frames: $W_p$ cached frames from recently generated chunks and $F$ live frames currently being denoised. Together with the cached state at position~0 and the live state at position~1, the visible window contains both clean and noisy entries at consistent noise levels: cached components at $\sigma{=}0$ and live components at the current denoising pass's noise level, matching the backbone's pretrained distribution without introducing any out-of-distribution asymmetry.

\begin{equation}
    \mathbf{h}_n = \big[\;\underbrace{\mathrm{KV}(s_{n-1})}_{F_s}\;;\; \underbrace{\tilde{s}_n}_{F_s}\;;\; \underbrace{\mathrm{KV}(\hat{\mathbf{x}}_c^{(\text{past})})}_{\text{$W_p$ frames}}\;;\; \underbrace{\mathbf{x}_c^{(n)}}_{\text{$F$ frames}}\;\big]
    \label{eq:window}
\end{equation}
 
\noindent\textbf{State positioning.} Under causal softmax, attention mass concentrates disproportionately on the earliest positions~\cite{xiao2024efficient}, making position~0 the most influential in the visible window. Existing methods exploit this by placing a frozen first frame there, which provides stable identity grounding but locks the scene reference for the entire generation. We instead place the adaptive state at this position, so that the model's dominant reference point evolves with the scene rather than remaining fixed. This choice also resolves a structural discontinuity introduced by block-relativistic RoPE~\cite{yesiltepe2025infinity}: while the sliding window is re-indexed at each chunk so that content positions are always relative, a static anchor at position~0 remains an absolute fixed point whose content never changes. An adaptive state that is regenerated each chunk makes position~0 consistent with the relative-time semantics of the rest of the window.
 
\noindent\textbf{Recurrence.} Both state and content start from independent Gaussian noise and are denoised jointly:
\begin{equation}
    [s_n,\, \hat{\mathbf{x}}_c^{(n)}] = G_\theta\!\left([\boldsymbol{\epsilon}_s,\, \boldsymbol{\epsilon}_c]\,;\; \mathbf{h}_n\right), \quad \boldsymbol{\epsilon}_s, \boldsymbol{\epsilon}_c \sim \mathcal{N}(0, I)
    \label{eq:denoise}
\end{equation}
where $\hat{\mathbf{x}}_c^{(n)}$ and $s_n$ are the clean content and state predictions, and $\mathbf{h}_n$ is the cache from Eq.~\eqref{eq:window}. After denoising, the two predictions follow different paths: the content is decoded into video and its clean KV enters the sliding window for short-term context, while the state is never decoded but instead overwrites position~0 to serve as the scene reference for the next chunk:
\begin{equation}
    \mathbf{h}^{(s)}_{n+1} = \mathrm{KV}(s_n), \qquad \mathrm{KV}(\hat{\mathbf{x}}_c^{(\text{past})}) = \mathrm{KV}(\hat{\mathbf{x}}_c^{(n-W_p+1 : n)})
    \label{eq:transition}
\end{equation}
This cache update creates the recurrence: when chunk $n{+}1$ begins, the generator denoises new content and state from noise, but the cache now carries $s_n$'s clean KV at position~0. Because the new state $s_{n+1}$ attends to this cached representation alongside the current content, the generator $G_\theta$ serves as the state transition function and the KV cache as the carrier. This parallels the recurrence in state models such as RNNs \cite{chung2014empirical} and LSTMs~\cite{hochreiter1997long}: a hidden variable, updated at each step by the model's own computation, that conditions the output without being observed directly.
 
\noindent\textbf{Information flow.} During each denoising pass, queries from the live tokens attend to the full visible window. This creates two complementary information flows: content queries read the cached state for scene context that has been evicted from the sliding window, while the live state's queries read current content to absorb the evolving scene. Because the live state starts from pure noise with no structural prior encoding the previous chunk, identity and scene context must be actively reconstructed through attention to the cached KV at position~0, rather than passively copied from the input. At chunk~0, no prior state exists; the first content frame's clean latent serves as $s_0$ and its KV initializes position~0. The state slot activates only when eviction begins, preserving the pretrained model's behavior when all content fits in the window.
\begin{equation}
    \mathrm{Attn}([\mathbf{q}_{\tilde{s}_n};\, \mathbf{q}_c^{\text{live}}],\; [\mathbf{k}_{s_{n-1}};\, \mathbf{k}_{\tilde{s}_n};\, \mathbf{k}_c^{\text{cache}};\, \mathbf{k}_c^{\text{live}}],\; [\mathbf{v}_{s_{n-1}};\, \mathbf{v}_{\tilde{s}_n};\, \mathbf{v}_c^{\text{cache}};\, \mathbf{v}_c^{\text{live}}])
    \label{eq:attn}
\end{equation}

Equation~\ref{eq:attn} makes explicit that the state participates in the same attention mechanism as content, no separate module or gating is required. The full recurrence thus consists of three standard operations: joint denoising (Eq.~\ref{eq:denoise}), cache update (Eq.~\ref{eq:transition}), and attention (Eq.~\ref{eq:attn}), all already present in the pretrained backbone. The state's representation is shaped entirely by the generation loss propagated through content attention; no auxiliary objective or supervision is needed to teach the model what to store.
 
\subsection{Horizon-Weighted Training}
\label{sec:training}
 
In autoregressive generation, errors propagate and amplify through the chunk chain, so the later training frames experience the most accumulated drift and preview beyond-horizon conditions. Under a uniform loss, these critical late frames are underweighted, early frames, being well-conditioned, dominate the mean loss and absorb optimizer capacity. The frames that matter most for generalization receive the least optimization pressure. We address this by weighting the DMD loss per frame with a linear ramp that increases with frame index:
\begin{equation}
    \mathcal{L} = \frac{1}{2}\sum_{i=0}^{N-1} w_i \left\|\hat{\mathbf{x}}_i - \left(\hat{\mathbf{x}}_i - \frac{S_\text{fake}(\hat{\mathbf{x}}_{i,t}, t) - S_\text{real}(\hat{\mathbf{x}}_{i,t}, t)}{\gamma}\right)_{\!\text{sg}}\right\|^2, \quad w_i = 1 + \alpha \cdot \frac{i}{N - 1}
    \label{eq:loss}
\end{equation}
where $\hat{\mathbf{x}}_i$ is the clean prediction of frame $i$, $\hat{\mathbf{x}}_{i,t} = (1{-}t)\hat{\mathbf{x}}_i + t\boldsymbol{\epsilon}$ is its noised version at level $t$, $N$ is the number of frames in the rollout, and $\alpha$ controls the ramp slope, redirecting the optimizer toward the later frames where drift accumulates. This weighting is particularly important for the adaptive state, since the late frames are precisely the ones where the original scene content has exited the sliding window and the cached state at position~0 becomes the primary scene reference. No separate loss is applied to the state; instead, gradient reaches it entirely through the attention that content frames pay to it. Because the horizon weighting concentrates the loss on the frames whose quality depends most on the state's contribution, the training signal naturally shapes the state to provide useful scene context where it matters most. The cross-chunk recurrence is detached at chunk boundaries to allow independent optimization of each chunk's state prediction.

\section{Experiments}
\label{sec:experiments}

\begin{figure}[t!]
    \centering
    \includegraphics[width=\linewidth]{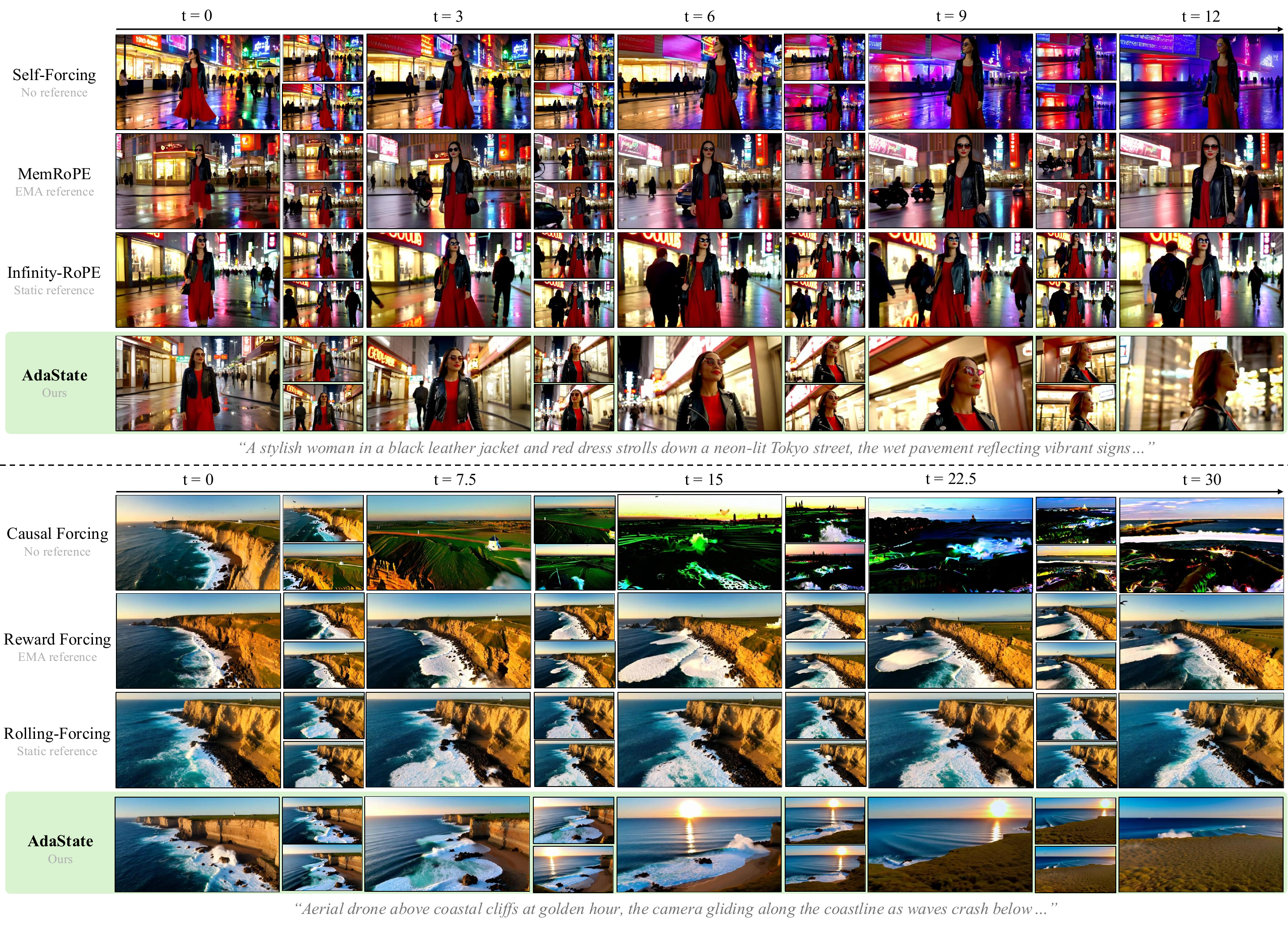}

    \caption{\textbf{Qualitative comparison across anchor categories.} Top block: 12-second generation sampled at 3-second intervals. Bottom block: 30-second generation sampled at 7.5-second intervals. Each block pairs AdaState against one exemplar per baseline category (no reference, EMA reference, static reference); the six exemplars across the two blocks cover all baseline groups. Methods without a persistent anchor accumulate color drift (Self-Forcing, top) or collapse to artifacts at long horizons (Causal Forcing, bottom). EMA and static references prevent collapse but freeze the scene: MemRoPE, Infinity-RoPE, Reward Forcing, and Rolling-Forcing each reproduce a near-identical composition across the entire rollout. AdaState alone preserves identity while producing natural scene evolution.}
    \label{fig:quali_comp}
    \vspace{-1em}
\end{figure}

We build on Wan2.1-T2V-1.3B~\cite{wan2025wan}, distilled into a causal autoregressive generator via Self-Forcing~\cite{huang2025selfforcing} with DMD loss against a Wan2.1-T2V-14B teacher. Each chunk denoises $F{=}3$ latent frames through a 4-step schedule alongside $F_s{=}1$ adaptive state frame, with $W_p{=}3$ cached content frames from prior chunks providing local context. Starting from the Self-Forcing checkpoint, we fine-tune on 21-frame rollouts (seven chunks) for 1000 iterations with horizon-weighted DMD using their training prompts, $\alpha{=}2$ for within-horizon evaluation, $\alpha{=}4$ for long-horizon generation, at learning rate $2{\times}10^{-6}$ and effective batch size 4 on two H200 GPUs. Evaluation uses VBench~\cite{huang2025vbench++} at 5 seconds (21 frames, within training horizon) and 30 seconds (120 frames, six times the training horizon), and VisionReward~\cite{xu2026visionreward} at 5 seconds; further details appear in the supplementary.
 
We compare against methods spanning anchor mechanisms for streaming generation: no persistent anchor (Self-Forcing~\cite{huang2025selfforcing}, CausVid~\cite{yin2025causvid}, Causal Forcing~\cite{zhu2026causal}), static anchor (LongLive~\cite{yang2026longlive}, Rolling Forcing~\cite{liu2025rolling}, Infinity-RoPE~\cite{yesiltepe2025infinity}), EMA and positional updates (Reward Forcing~\cite{lu2025reward}, MemRoPE~\cite{kim2026memrope}), and a heuristic anchor replacement (Rolling Sink~\cite{li2026rolling}). The non-autoregressive Wan 2.1-1.3B serves as a non-autoregressive quality reference.

\subsection{Qualitative Results}
Figure~\ref{fig:quali_comp} pairs AdaState against one exemplar per baseline category at two horizons. The top block presents a 12-second portrait rollout. Self-Forcing, trained on 5-second rollouts, accumulates visual artifacts beyond its training horizon: color drift emerges early and compounds through the rest of the sequence. MemRoPE and Infinity-RoPE prevent drift but lock the composition: their rollouts reproduce a near-identical scene across all keyframes, with no progression of camera or scene. The bottom block extends to 30 seconds on a drone shot of a coastline at golden hour. Causal Forcing, similarly trained on 5-second rollouts, collapses to visual artifacts well beyond its training horizon, no longer corresponding to the prompt. Rolling-Forcing and Reward Forcing prevent the collapse but freeze the scene similarly, with their static/EMA references. AdaState alone produces both temporal stability and natural progression: across the 12-second portrait shot, the scene evolves continuously with camera motion and subject action; across the 30-second coastal drone shot, the camera glides along the shoreline, revealing new terrain in continuous golden hour light. We provide further qualitative examples in the supplementary material.

\subsection{Quantitative Results}
\label{sec:main_results}

\begin{table*}[t]
\centering
\caption{\textbf{Evaluation in 5s and 30s rollouts.} 5 seconds (21 frames) is within all methods' training horizons; 30 seconds (120 frames) is six times the training length. VisionReward (VR) is reported at 5 seconds only. Best in \textbf{bold}, second best \underline{underlined}.}
\label{tab:vbench}
\resizebox{\textwidth}{!}{%
\begin{tabular}{l ccc >{\columncolor{blue!6}}c cc ccc >{\columncolor{blue!6}}c c}
\toprule
& \multicolumn{6}{c}{\textbf{5 seconds (within training horizon)}} & \multicolumn{5}{c}{\textbf{30 seconds (6$\times$ training horizon)}} \\
\cmidrule(lr){2-7} \cmidrule(lr){8-12}
\textbf{Method} & Subj.$\uparrow$ & Dyn.$\uparrow$ & Aesth.$\uparrow$ & Total$\uparrow$ & Consist.$\uparrow$ & VR$_{(\times 10)}\!\uparrow$ & Subj.$\uparrow$ & Dyn.$\uparrow$ & Aesth.$\uparrow$ & Total$\uparrow$ & Consist.$\uparrow$ \\
\midrule
\rowcolor{gray!8}
\multicolumn{12}{l}{\textit{Non-autoregressive}} \\
Wan 2.1-1.3B            & 0.970 & 0.609 & 0.593 & 0.823 & 0.253 & 0.664 & --    & --    & --    & --    & --    \\
\midrule
\rowcolor{gray!8}
\multicolumn{12}{l}{\textit{No persistent anchor}} \\
Self-Forcing            & 0.966 & 0.635 & 0.619 & 0.832 & 0.249 & 0.806 & 0.972 & 0.392 & 0.570 & 0.794 & 0.224 \\
CausVid                 & 0.979 & 0.547 & 0.606 & 0.819 & 0.238 & 0.620 & \underline{0.986} & 0.096 & 0.626 & 0.761 & 0.213 \\
Causal Forcing          & 0.974 & \underline{0.747} & 0.628 & \underline{0.851} & 0.246 & 0.774 & 0.944 & \textbf{0.933} & 0.523 & 0.843 & 0.194 \\
\midrule
\rowcolor{gray!8}
\multicolumn{12}{l}{\textit{Static anchor}} \\
LongLive                & 0.976 & 0.393 & \textbf{0.635} & 0.806 & 0.253 & 0.840 & 0.983 & 0.373 & 0.625 & 0.801 & \underline{0.255} \\
Rolling Forcing         & \textbf{0.983} & 0.367 & 0.619 & 0.804 & 0.251 & \underline{0.864} & \textbf{0.991} & 0.042 & \textbf{0.648} & 0.769 & 0.254 \\
Infinity-RoPE           & 0.978 & 0.628 & 0.605 & 0.832 & 0.245 & 0.820 & 0.986 & 0.267 & \underline{0.638} & 0.788 & 0.237 \\
\midrule
\rowcolor{gray!8}
\multicolumn{12}{l}{\textit{EMA / positional / heuristic}} \\
Reward Forcing          & 0.975 & 0.643 & 0.622 & 0.838 & 0.248 & 0.791 & 0.979 & 0.607 & 0.628 & 0.837 & 0.252 \\
MemRoPE                 & 0.976 & 0.622 & 0.622 & 0.836 & 0.253 & 0.798 & 0.974 & 0.718 & 0.601 & \underline{0.848} & 0.249 \\
Rolling Sink            & \underline{0.979} & 0.508 & 0.627 & 0.822 & \textbf{0.255} & 0.812 & 0.977 & 0.402 & 0.621 & 0.807 & 0.252 \\
\midrule
\rowcolor{gray!8}
\multicolumn{12}{l}{\textit{Adaptive anchor (ours)}} \\
\textbf{AdaState}       & 0.961 & \textbf{0.828} & \underline{0.630} & \textbf{0.860} & \underline{0.254} & \textbf{0.868} & 0.959 & \underline{0.922} & 0.596 & \textbf{0.865} & \textbf{0.259} \\
\bottomrule
\end{tabular}%
}
\end{table*}

\begin{wrapfigure}{r}{0.5\linewidth}
    \centering
    \includegraphics[width=\linewidth]{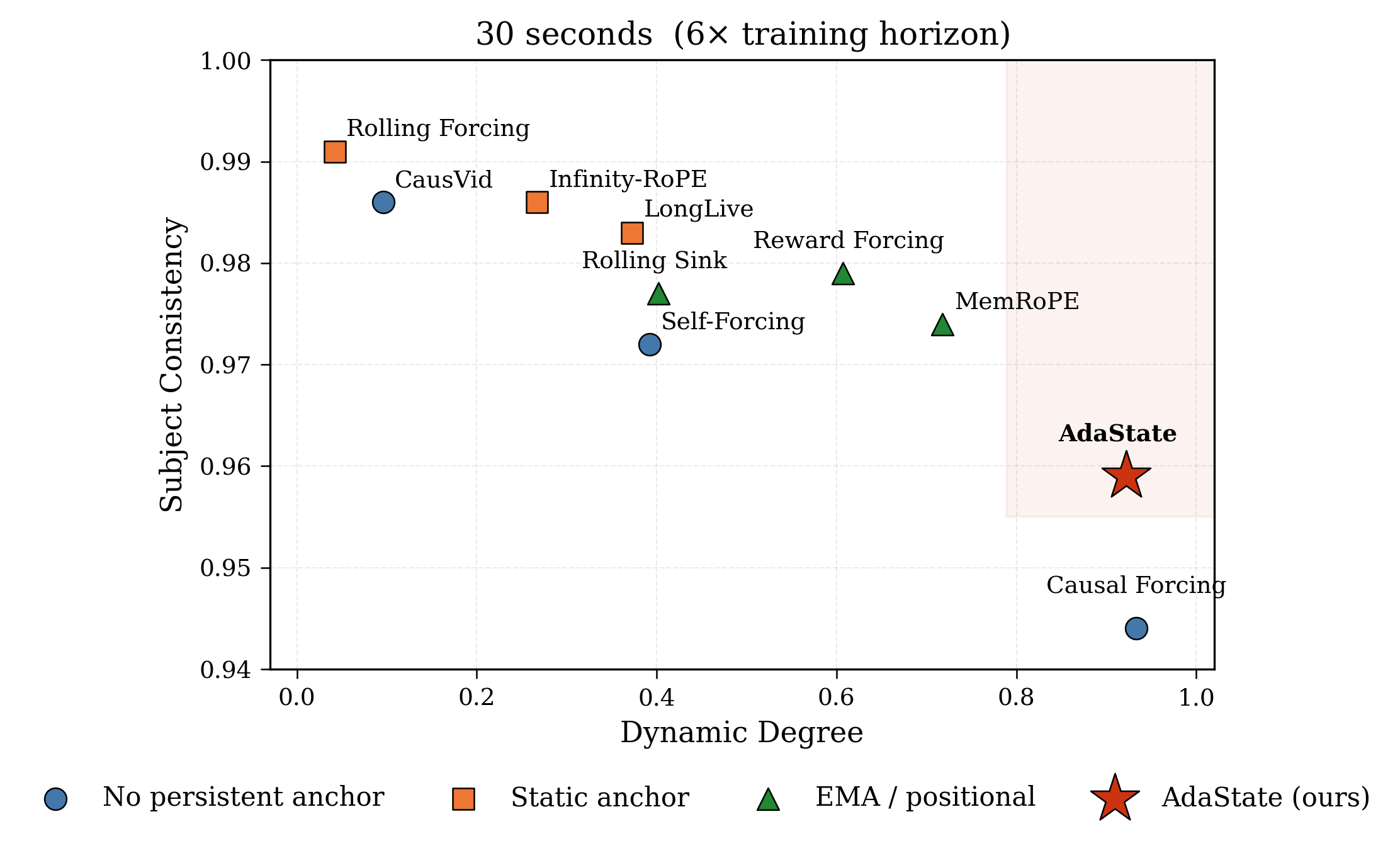}

    \caption{\textbf{Subject consistency vs.\ dynamic degree at 30 seconds.} The dynamics distribution becomes bimodal: most baselines collapse to the left as motion stops, while AdaState alone occupies the shaded upper-right region where high dynamics and high consistency coexist.}
    \label{fig:scatter}
    \vspace{-1em}
\end{wrapfigure}

We frame streaming methods by their anchor mechanism: no persistent anchor, static reference, EMA-based or heuristic update, and adaptive. The first three categories share a structural limit: content-agnostic anchor handling forces a consistency-dynamics tradeoff, while only the adaptive design can break it. We tested this behavior at two horizons: within (5 seconds) and beyond training (30 seconds, six times the training horizon). Evaluation uses 128 prompts from MovieGenBench\cite{polyak2024movie}: a mix of static and dynamic scenes at 5 seconds, and prompts implying moving scenes at 30 seconds to further stress-test long-horizon dynamics. We report VBench~\cite{huang2025vbench++} and VisionReward~\cite{xu2026visionreward} in Table~\ref{tab:vbench}, with the consistency-dynamics tradeoff visualized at long horizon in Figure~\ref{fig:scatter}. Within the training horizon, baselines populate the predicted diagonal by category: static-anchor methods (LongLive, Rolling Forcing) cluster at the high-consistency, low-dynamics end; anchor-free methods (Self-Forcing, Causal Forcing) sit at the opposite end, recovering dynamics at the cost of consistency; EMA and heuristic anchor-update methods land along the same line without escaping it, since content-agnostic updates cannot adapt the reference to evolving content. AdaState is the only method off the diagonal: it posts the highest VBench total, dynamic degree, and VisionReward, with subject consistency just below the static-anchor cluster, the price of replacing a frozen reference with an evolving one.

We then push to 30 seconds to stress-test whether the tradeoff holds beyond training horizon. The categorical structure persists and sharpens (Figure~\ref{fig:scatter}): six of nine baselines fall below 0.5 dynamic degree with Rolling Forcing and CausVid effectively frozen, and among methods that maintain motion, Causal Forcing pays by drifting off the prompt, its subject consistency dropping noticeably from the 5-second value. AdaState alone occupies the upper-right corner, preserving dynamics without losing consistency, and posts the highest VBench total and text-video alignment at this horizon. Crucially, AdaState's dynamic degree climbs from 5 to 30 seconds rather than degrading, and the gap to Self-Forcing widens across the same range, direct evidence that horizon-weighted DMD concentrates gradient on the late-rollout regime where state-mediated coherence matters most.

\begin{wrapfigure}{r}{0.5\linewidth}
    \centering
    \includegraphics[width=\linewidth]{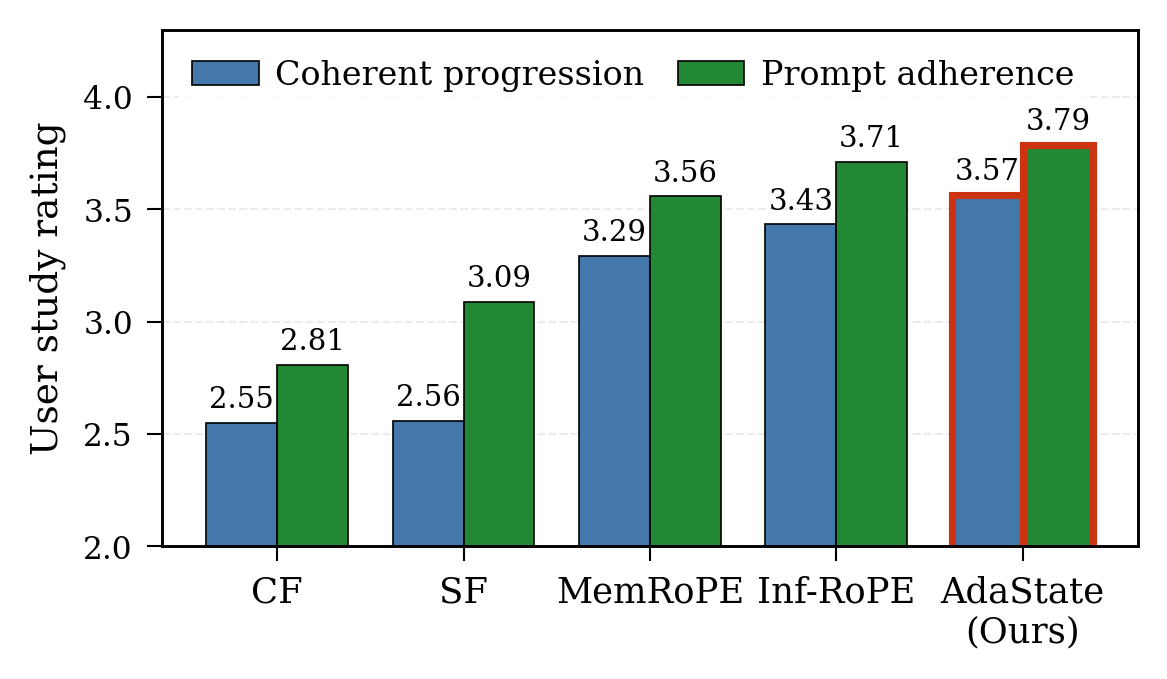}

    \caption{\textbf{User study (5-point Likert, $N{=}40$ raters).} Methods are ordered by coherent-progression score. AdaState gets the highest ratings on both coherent progression and prompt following.}
    \label{fig:user_study}
\end{wrapfigure}

To confirm the perceptual ranking, we conduct a user study with 40 raters with Prolific platform\footnote{\url{https://www.prolific.com/}}. Each rater views videos from AdaState and four representative baselines: Self-Forcing and Causal Forcing (no-reference), Infinity-RoPE (static-reference), and MemRoPE (EMA-reference), on 20 videos spanning both horizons, scoring each on coherent progression and prompt adherence. The ratings mirror the categorical structure (Figure~\ref{fig:user_study}): no-reference methods score lowest, static and EMA-reference methods occupy the middle, and AdaState leads on both dimensions. The largest gap falls against the no-reference category, where beyond-horizon artifacts penalize both ratings; the ordering matches the failure modes from Figure~\ref{fig:quali_comp}, with locked or artifact-laden videos rated below those that progress coherently. We provide further details on our user study in the supplementary material.

\subsection{Ablation Study}
\label{sec:ablations}

\begin{table}[t]
\centering
\caption{\textbf{Ablation study at 5s (21 frames).} We vary the state size $F_s$, content window $W_p{+}F$, loss weighting $\alpha$, and anchor configuration. $\uparrow$ = higher is better.}
\label{tab:ablation}
\resizebox{\columnwidth}{!}{%
\begin{tabular}{lcccc ccccccc}
\toprule
\textbf{Configuration} & \textbf{Sink} & $F_s$ & $W_p{+}F$ & $\alpha$ & Subj.$\uparrow$ & BG$\uparrow$ & Smooth$\uparrow$ & Dyn.$\uparrow$ & Aesth.$\uparrow$ & Img.$\uparrow$ & Avg.$\uparrow$ \\
\midrule
State, $F_s{=}3$             & --         & 3 & 9 & --  & 0.971 & 0.964 & 0.984 & 0.602 & 0.623 & \textbf{0.721} & 0.811 \\
State, $F_s{=}1$             & --         & 1 & 9 & --  & 0.962 & 0.959 & 0.980 & 0.581 & 0.631 & 0.700 & 0.802 \\
State, $F_s{=}1$             & --         & 1 & 6 & --  & 0.965 & 0.956 & 0.981 & 0.734 & 0.627 & 0.707 & 0.829 \\
\midrule
State, $F_s{=}1$, $\alpha{=}2$ & --       & 1 & 6 & 2   & 0.961 & 0.953 & 0.980 & \textbf{0.828} & \textbf{0.630} & 0.702 & \textbf{0.842} \\
State, $F_s{=}1$, $\alpha{=}4$ & --       & 1 & 6 & 4   & 0.971 & 0.961 & 0.984 & 0.688 & 0.628 & 0.712 & 0.824 \\
\midrule
Sink + State, $\alpha{=}4$   & \checkmark & 1 & 6 & 4   & \textbf{0.977} & \textbf{0.966} & \textbf{0.986} & 0.570 & 0.629 & 0.703 & 0.805 \\
\bottomrule
\end{tabular}%
}
\end{table}

We ablate three design choices (Table~\ref{tab:ablation}): the state architecture, horizon weighting $\alpha$, and whether the adaptive state coexists with a retained static anchor. The architecture ablation reveals a proportional design rule: dynamics depend on the state's relative share within the cache window, not on the absolute capacity of either component. Holding $F_s{=}1$ and shrinking the content window from 9 to 6 frames raises dynamics substantially because the state's share grows while staying concentrated at a single recurrent position. Expanding to $F_s{=}3$ with the larger window allocates more raw capacity to state but spreads it across multiple slots, diluting the singular anchor effect; dynamics do not recover. A single state slot in a tight content window concentrates attention mass at one recurrent position more effectively than multiple slots in a wider one.

The horizon weighting ablation motivates our two-regime training. At $\alpha{=}2$, dynamics and total score peak, the right choice within the training horizon. At $\alpha{=}4$, the optimizer concentrates more gradient on late frames, trading within-horizon dynamic range for higher consistency and imaging quality at the rollout's end. We therefore use $\alpha{=}2$ for within-horizon generation and $\alpha{=}4$ for long-horizon generation, where the late-frame quality directly benefits the extended rollout. We provide further ablations in the supplementary material.

Finally, retaining a static sink alongside the adaptive state suppresses dynamics. The model shortcuts through the clean static reference even when an evolving alternative is available, confirming that removing the static anchor is necessary for the adaptive state to function.

\section{Limitations} 
The state's capacity is set heuristically at a single frame ($F_s{=}1$). Our ablations show that adding more state slots at the current training horizon does not improve dynamics; the single slot already carries the scene context that motion requires within the rollouts we evaluate. For longer rollouts or more complex scenes (multiple actors, concurrent events, parallel scene structure), explicit memory mechanisms that augment the adaptive state with structured external memory may become necessary, and we leave this combination as future work. More broadly, AdaState proposes a recurrent formulation for propagating information across steps in autoregressive generation, a hidden state generated alongside content and propagated through cached attention, that should apply wherever attention concentrates at a fixed past position and a recency window, the attention-sink pattern characterized broadly in transformers~\cite{xiao2024efficient}. We investigate this formulation in video, where the consistency-dynamics tradeoff most clearly motivates the design, and the construction opens a path to adaptive recurrent state in autoregressive generation more broadly.

\section{Conclusion} 
\label{sec:conclusion}
Streaming video generation faces a structural tradeoff: the position-zero anchor that enables long-horizon stability also determines within-horizon scene progression. AdaState replaces this static anchor with a hidden latent that is denoised alongside content at every chunk but never rendered, propagated across chunks through its cached KV in the same slot, and trained with horizon-weighted DMD so the anchor evolves in service of motion rather than fighting it. The construction requires no architectural change, no auxiliary loss, and no external memory. Across both within- and beyond-horizon evaluation, AdaState breaks the consistency-dynamics tradeoff that other anchor mechanisms fall along, and the gap to baselines widens with rollout length. A user study confirms the perceptual ranking on both coherent progression and prompt adherence.

\section*{Acknowledgements}

We gratefully acknowledge \href{https://fal.ai}{fal.ai} for providing compute support through the project. Yusuf Dalva is supported by a grant from
the Amazon-Virginia Tech Initiative for Efficient and Robust Machine
Learning. This research is supported by the National Science Foundation
under Grant No.\ 2543524.

\bibliographystyle{plainnat}
\bibliography{references}

\newpage
\appendix

\section*{Table of Contents}
\addcontentsline{toc}{section}{Supplementary Material Table of Contents}
\startcontents[appendix]
\printcontents[appendix]{l}{1}{\setcounter{tocdepth}{2}}

\newpage

\section{Implementation Details}
\label{app:implementation}
 
This section provides the full training algorithm and hyperparameter configuration for AdaState. We build on the Self-Forcing codebase and checkpoint, modifying only the cache management (state slot at position~0, content sliding window) and the loss weighting (horizon ramp). No architectural changes are made to the backbone DiT; the adaptive state occupies the same latent space as content frames and is processed by the same layers.
 
\paragraph{Training algorithm.}
Algorithm~\ref{alg:training} summarizes the AdaState training procedure. Each iteration performs a full autoregressive rollout of $N_c$ chunks. At each chunk, the generator jointly denoises $F_s$ state frames and $F$ content frames from independent Gaussian noise, conditioned on the KV cache. After denoising, the content's clean KV enters the sliding window while the state's clean KV overwrites position~0. The horizon-weighted DMD loss (Eq.~5 in the main paper) is applied per frame, and the cross-chunk state recurrence is detached at chunk boundaries.
 
\begin{algorithm}[ht]
\caption{AdaState Training}
\label{alg:training}
\begin{algorithmic}[1]
\Require Generator $G_\theta$, critic $S_\text{fake}$, frozen teacher $S_\text{real}$, chunk count $N_c$, chunk size $F$, state size $F_s$, cached content frames $W_p$, denoising steps $K$, horizon weight $\alpha$, learning rate $\eta$
\While{training}
    \State Sample prompt $p$
    \State Generate first chunk $\mathbf{x}_c^{(0)}$ from teacher, $\;s_0 \gets \mathbf{x}_c^{(0)}[0]$ \Comment{init state from first frame}
    \State $\mathbf{h} \gets \mathrm{KV}(s_0) \;\|\; \mathrm{KV}(\mathbf{x}_c^{(0)})$ \Comment{init cache}
    \For{$n = 1, \cdots, N_c - 1$}
        \State Sample $\boldsymbol{\epsilon}_s, \boldsymbol{\epsilon}_c \sim \mathcal{N}(0, I)$
        \State $[s_n,\, \hat{\mathbf{x}}_c^{(n)}] \gets G_\theta([\boldsymbol{\epsilon}_s, \boldsymbol{\epsilon}_c]\,;\; \mathbf{h})$ \Comment{joint state + content denoising}
        \State $w_i = 1 + \alpha \cdot i\,/\,(N{-}1)$ for global frame index $i$ \Comment{horizon weights}
        \State $\hat{\mathbf{x}}_{i,t} = (1{-}t)\,\hat{\mathbf{x}}_i + t\,\boldsymbol{\epsilon}, \quad t \sim \mathcal{U}(0, 1)$ \Comment{noise predictions}
        \State $\nabla_\text{DMD} = (S_\text{fake}(\hat{\mathbf{x}}_{i,t}, t) - S_\text{real}(\hat{\mathbf{x}}_{i,t}, t))\,/\,\gamma$ \Comment{DMD gradient}
        \State $\mathcal{L}_n = \frac{1}{2}\sum_{i} w_i \left\| \hat{\mathbf{x}}_i - \left(\hat{\mathbf{x}}_i - \nabla_\text{DMD}\right)_{\text{sg}} \right\|^2$ \Comment{weighted loss}
        \State $\mathrm{KV}(\hat{\mathbf{x}}_c^{(\text{past})}) \gets \mathrm{KV}(\hat{\mathbf{x}}_c^{(n-W_p+1:n)})$ \Comment{slide content window}
        \State $\mathbf{h}^{(s)} \gets \mathrm{KV}(\text{stopgrad}(s_n))$ \Comment{detach state across chunks}
    \EndFor
    \State $\mathcal{L} \gets \sum_{n} \mathcal{L}_n$
    \State $\theta \gets \theta - \eta\,\nabla_\theta \mathcal{L}$ \Comment{generator update}
    \State Update $S_\text{fake}$ on $\{(\hat{\mathbf{x}}_{i,t}, t)\}$ \Comment{critic update}
\EndWhile
\end{algorithmic}
\end{algorithm}
 
\paragraph{Hyperparameters.}
Table~\ref{tab:hparams} lists the full training configuration. We fine-tune from the Self-Forcing checkpoint for 1000 iterations using AdamW ($\beta_1{=}0.9$, $\beta_2{=}0.999$, weight decay $0.01$) on two H200 GPUs. The only hyperparameter that differs between models is the horizon weight $\alpha$: we use $\alpha{=}2$ for within-horizon evaluation and $\alpha{=}4$ for long-horizon generation, where concentrating gradient on late frames directly benefits the extended rollout. EMA averaging begins at iteration 200 with decay 0.99.
 
\begin{table}[h]
\centering
\small
\caption{Training hyperparameters.}
\label{tab:hparams}
\resizebox{\linewidth}{!}{%
\begin{tabular}{@{}cccccccccccc@{}}
\toprule
Gen. LR & Critic LR & Optimizer & Batch size & Hardware & Chunk size ($F$) & State size ($F_s$) & Sliding window ($W_p$) & Denoise steps ($K$) & Rollout & Iterations \\
\midrule
$2{\times}10^{-6}$ & $4{\times}10^{-7}$ & AdamW & 4 & 2$\times$H200 & 3 frames & 1 frame & 3 frames & 4 & 21 frames / 7 chunks & 1000 \\
\bottomrule
\end{tabular}}
\end{table}

\section{User Study Details}
\label{app:userstudy}
 
We conduct a user study to evaluate whether human perception aligns with the categorical structure observed in automated metrics. The study compares AdaState against four representative baselines spanning the three baseline categories: Self-Forcing \cite{huang2025selfforcing} and Causal Forcing \cite{zhu2026causal} (no persistent anchor), Infinity-RoPE \cite{yesiltepe2025infinity} (static anchor), and MemRoPE \cite{kim2026memrope} (EMA-based anchor).

\begin{figure}[h]
    \centering
    \includegraphics[width=0.85\linewidth]{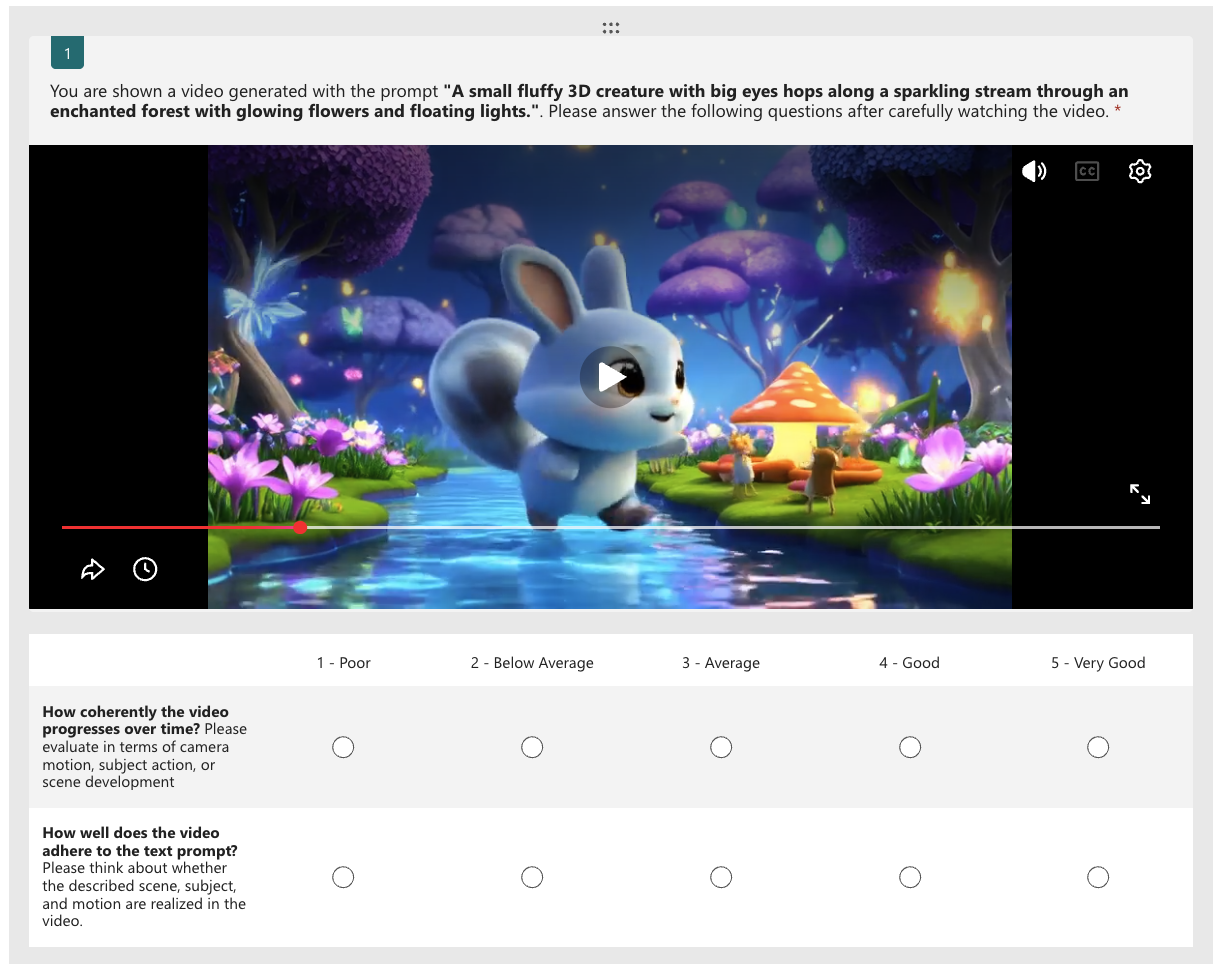}
    \caption{\textbf{User study evaluation interface. }Each rater views a video generated from a given prompt and scores it on two dimensions using a 5-point Likert scale. Method identity is hidden; video order is randomized. Identifying information has been redacted for anonymity.}
    \label{fig:userstudy_interface}
\end{figure}

We recruit 40 raters through the Prolific platform. Each rater evaluates 20 distinct videos, presented in randomized order with method identity hidden. The prompt set includes 10 within-horizon (5-second) and 10 beyond-horizon (30-second) generations to test whether perceptual quality degrades with rollout length. Videos are hosted as unlisted uploads (anonymous) and embedded in a Microsoft Forms questionnaire; raters are instructed to watch each video in full before responding. Figure~\ref{fig:userstudy_interface} shows the evaluation interface.
 
\paragraph{Rating dimensions.}
Each video is scored on two dimensions using a 5-point Likert scale (1 = Poor, 5 = Very Good):
\begin{itemize}
    \item \textbf{Coherent progression:} ``How coherently the video progresses over time? Please evaluate in terms of camera motion, subject action, or scene development.''
    \item \textbf{Prompt adherence:} ``How well does the video adhere to the text prompt? Please think about whether the described scene, subject, and motion are realized in the video.''
\end{itemize}
 
\section{Evaluation Details}
\label{app:vbench}
 
\paragraph{Detailed Quantitative Results.} Tables~\ref{tab:vbench5s} and~\ref{tab:vbench30s} report the per-dimension VBench scores at 5 and 30 seconds, extending Table~1 of the main paper. We report all ten dimensions: subject consistency, background consistency, motion smoothness, dynamic degree, aesthetic quality, imaging quality, temporal flickering, VBench total, overall consistency (text-video alignment), and VisionReward (5s only). Note that dynamic degree is a binary per-clip metric (1 if the scene contains meaningful motion, 0 otherwise), averaged across prompts; its standard deviation is inherently high because it reflects a binary decision rather than a continuous score. The categorical structure described in Section~4.2 holds across all dimensions: static-anchor methods dominate consistency and flickering metrics at the cost of dynamics, while anchor-free methods show the opposite pattern. AdaState achieves the highest total score at both horizons without trading off any single dimension catastrophically, its lowest relative standing is on temporal flickering and subject consistency, the expected cost of replacing a frozen anchor with an evolving one.

\newcommand{\eb}{{\scriptsize\,$\pm$\,.}}
\begin{table}[h]
\centering
\small
\caption{Full VBench breakdown at 5 seconds (21 frames). VisionReward (VR) is reported $\times 10$. Best in \textbf{bold}, second best \underline{underlined}.}
\label{tab:vbench5s}
\resizebox{\linewidth}{!}{%
\begin{tabular}{@{}l ccccccc >{\columncolor{blue!6}}c c c@{}}
\toprule
\textbf{Method} & Subj.$\uparrow$ & BG$\uparrow$ & Smooth$\uparrow$ & Dyn.$\uparrow$ & Aesth.$\uparrow$ & Img.$\uparrow$ & Flicker$\uparrow$ & Total$\uparrow$ & Consist.$\uparrow$ & VR$_{(\times 10)}\!\uparrow$ \\
\midrule
\rowcolor{gray!8}
\multicolumn{11}{l}{\textit{Non-autoregressive}} \\
Wan 2.1-1.3B & 0.970\eb{03} & 0.965\eb{01} & 0.986\eb{01} & 0.609\eb{43} & 0.593\eb{09} & 0.661\eb{10} & 0.976\eb{02} & 0.823 & 0.253\eb{05} & 0.664\eb{07} \\
\midrule
\rowcolor{gray!8}
\multicolumn{11}{l}{\textit{No persistent anchor}} \\
Self-Forcing & 0.966\eb{03} & 0.952\eb{02} & 0.983\eb{01} & 0.635\eb{44} & 0.619\eb{09} & 0.697\eb{07} & 0.968\eb{03} & 0.832 & 0.249\eb{05} & 0.806\eb{06} \\
CausVid & 0.979\eb{02} & 0.967\eb{01} & 0.980\eb{01} & 0.547\eb{45} & 0.606\eb{10} & 0.689\eb{08} & 0.968\eb{02} & 0.819 & 0.238\eb{05} & 0.620\eb{06} \\
Causal Forcing & 0.974\eb{02} & 0.965\eb{01} & 0.976\eb{02} & \underline{0.747}\eb{39} & 0.628\eb{09} & 0.710\eb{07} & 0.955\eb{03} & \underline{0.851} & 0.246\eb{05} & 0.774\eb{06} \\
\midrule
\rowcolor{gray!8}
\multicolumn{11}{l}{\textit{Static anchor}} \\
LongLive & 0.976\eb{02} & 0.967\eb{02} & 0.989\eb{01} & 0.393\eb{44} & \underline{0.635}\eb{09} & 0.700\eb{07} & \textbf{0.998}\eb{02} & 0.806 & 0.253\eb{05} & 0.840\eb{06} \\
Rolling Forcing & \underline{0.983}\eb{01} & 0.971\eb{01} & 0.988\eb{01} & 0.367\eb{43} & 0.619\eb{09} & \underline{0.721}\eb{05} & 0.978\eb{01} & 0.804 & 0.251\eb{05} & \underline{0.864}\eb{09} \\
Infinity-RoPE & 0.978\eb{02} & 0.968\eb{01} & 0.986\eb{01} & 0.628\eb{43} & 0.605\eb{09} & 0.683\eb{08} & 0.975\eb{02} & 0.832 & 0.245\eb{05} & 0.820\eb{06} \\
\midrule
\rowcolor{gray!8}
\multicolumn{11}{l}{\textit{EMA / positional / heuristic}} \\
Reward Forcing & 0.975\eb{02} & 0.967\eb{01} & 0.984\eb{01} & 0.643\eb{46} & 0.622\eb{09} & 0.704\eb{07} & 0.969\eb{02} & 0.838 & 0.248\eb{05} & 0.791\eb{07} \\
MemRoPE & 0.976\eb{02} & 0.967\eb{01} & 0.983\eb{01} & 0.622\eb{43} & 0.622\eb{09} & 0.712\eb{07} & 0.968\eb{01} & 0.836 & \underline{0.253}\eb{05} & 0.798\eb{06} \\
Rolling Sink & 0.979\eb{02} & \underline{0.970}\eb{01} & 0.986\eb{01} & 0.508\eb{44} & 0.627\eb{09} & 0.708\eb{07} & 0.975\eb{01} & 0.822 & \textbf{0.255}\eb{03} & 0.812\eb{06} \\
\midrule
\rowcolor{gray!8}
\multicolumn{11}{l}{\textit{Adaptive anchor (ours)}} \\
\textbf{AdaState} & 0.961\eb{03} & 0.953\eb{02} & 0.980\eb{02} & \textbf{0.828}\eb{39} & \textbf{0.630}\eb{09} & 0.702\eb{07} & 0.963\eb{02} & \textbf{0.860} & 0.254\eb{05} & \textbf{0.868}\eb{05} \\
\bottomrule
\end{tabular}}
\end{table}
 
\begin{table}[h]
\centering
\small
\caption{Full VBench breakdown at 30 seconds (120 frames, 6$\times$ training horizon). Best in \textbf{bold}, second best \underline{underlined}.}
\label{tab:vbench30s}
\resizebox{\linewidth}{!}{%
\begin{tabular}{@{}l ccccccc >{\columncolor{blue!6}}c c@{}}
\toprule
\textbf{Method} & Subj.$\uparrow$ & BG$\uparrow$ & Smooth$\uparrow$ & Dyn.$\uparrow$ & Aesth.$\uparrow$ & Img.$\uparrow$ & Flicker$\uparrow$ & Total$\uparrow$ & Consist.$\uparrow$ \\
\midrule
\rowcolor{gray!8}
\multicolumn{10}{l}{\textit{No persistent anchor}} \\
Self-Forcing & 0.972\eb{01} & 0.962\eb{01} & 0.991\eb{01} & 0.392\eb{31} & 0.570\eb{05} & 0.686\eb{08} & 0.983\eb{01} & 0.794 & 0.224\eb{05} \\
CausVid & \underline{0.986}\eb{00} & 0.972\eb{01} & \underline{0.991}\eb{00} & 0.096\eb{12} & 0.626\eb{10} & 0.666\eb{10} & \underline{0.987}\eb{01} & 0.761 & 0.213\eb{04} \\
Causal Forcing & 0.944\eb{01} & 0.949\eb{01} & 0.970\eb{02} & \underline{0.933}\eb{11} & 0.523\eb{05} & 0.637\eb{09} & 0.944\eb{03} & \underline{0.843} & 0.194\eb{03} \\
\midrule
\rowcolor{gray!8}
\multicolumn{10}{l}{\textit{Static anchor}} \\
LongLive & 0.983\eb{01} & 0.967\eb{01} & 0.990\eb{00} & 0.373\eb{34} & 0.625\eb{04} & 0.692\eb{07} & 0.978\eb{01} & 0.801 & \underline{0.255}\eb{03} \\
Rolling Forcing & \textbf{0.991}\eb{00} & \textbf{0.978}\eb{01} & \textbf{0.992}\eb{00} & 0.042\eb{15} & \underline{0.648}\eb{04} & \textbf{0.747}\eb{06} & \textbf{0.986}\eb{01} & 0.769 & 0.254\eb{03} \\
Infinity-RoPE & 0.986\eb{01} & 0.972\eb{01} & 0.991\eb{01} & 0.267\eb{33} & 0.637\eb{05} & 0.680\eb{11} & 0.983\eb{01} & 0.788 & 0.237\eb{03} \\
\midrule
\rowcolor{gray!8}
\multicolumn{10}{l}{\textit{EMA / positional / heuristic}} \\
Reward Forcing & 0.979\eb{01} & 0.967\eb{01} & 0.989\eb{01} & 0.607\eb{40} & 0.628\eb{05} & 0.720\eb{08} & 0.970\eb{01} & 0.837 & 0.252\eb{03} \\
MemRoPE & 0.974\eb{01} & 0.962\eb{01} & 0.990\eb{00} & 0.718\eb{33} & 0.601\eb{04} & \underline{0.717}\eb{08} & 0.973\eb{01} & 0.848 & 0.249\eb{03} \\
Rolling Sink & 0.977\eb{01} & 0.965\eb{01} & 0.990\eb{01} & 0.402\eb{37} & 0.621\eb{04} & 0.711\eb{08} & 0.980\eb{01} & 0.807 & 0.252\eb{03} \\
\midrule
\rowcolor{gray!8}
\multicolumn{10}{l}{\textit{Adaptive anchor (ours)}} \\
\textbf{AdaState} & 0.959\eb{01} & 0.951\eb{01} & 0.980\eb{01} & \textbf{0.922}\eb{10} & 0.596\eb{04} & 0.685\eb{09} & 0.962\eb{02} & \textbf{0.865} & \textbf{0.259}\eb{03} \\
\bottomrule
\end{tabular}}
\end{table}

\paragraph{Evaluation prompts.}
We use two prompt sets: general-purpose prompts from MovieGenBench Extended (provided by Self-Forcing) for 5-second evaluation, covering a mix of static and dynamic scenes across diverse subjects and styles; and prompts for 30-second evaluation, all describing aerial or drone-style moving camera shots over varied landscapes and cityscapes, designed to stress-test long-horizon dynamics. Tables~\ref{tab:prompts5s} and~\ref{tab:prompts30s} show representative examples from each set.
 
\begin{table}[h]
\centering
\small
\caption{Representative prompts from the 5-second evaluation set.}
\label{tab:prompts5s}
\begin{tabular}{@{}p{0.95\linewidth}@{}}
\toprule
\textbf{Prompt} \\
\midrule
A stylish woman strolls down a bustling Tokyo street, the warm glow of neon lights and animated city signs casting vibrant reflections. She wears a sleek black leather jacket paired with a flowing red dress and black boots\ldots \\
\midrule
A 3D animation of a small, round, fluffy creature with big, expressive eyes exploring a vibrant, enchanted forest. The creature, a whimsical blend of a rabbit and a squirrel, has soft blue fur and a bushy, striped tail\ldots \\
\midrule
A winter scene in a snowy forest, where a litter of playful golden retriever puppies emerge from the snow. Their heads pop out, their fluffy fur glistening in the sunlight, and they wag their tails joyfully\ldots \\
\midrule
A dramatic and dynamic moment captured in a realistic photographic style, featuring a golden retriever dog leaping into a pool to rescue a child. The dog is mid-jump, its legs stretched forward and its fur glistening in the sunlight\ldots \\
\midrule
A scenic photograph capturing the moment a steam train departs from the Glenfinnan Viaduct, a historic railway bridge in Scotland. The train moves gracefully over the arch-covered viaduct, its smoke billowing into the air\ldots \\
\bottomrule
\end{tabular}
\end{table}
 
\begin{table}[h]
\centering
\small
\caption{Representative prompts from the 30-second evaluation set.}
\label{tab:prompts30s}
\begin{tabular}{@{}p{0.95\linewidth}@{}}
\toprule
\textbf{Prompt} \\
\midrule
A sweeping aerial drone shot above dramatic coastal cliffs at golden hour, deep blue waves crashing against the rocks below and sending up white spray. The camera glides along the coastline at altitude, revealing rugged stone faces\ldots \\
\midrule
A dynamic FPV aerial flight through a sprawling cyberpunk megacity at night, neon billboards and holographic advertisements glowing pink, blue, and violet. The camera weaves between towering skyscrapers\ldots \\
\midrule
A dynamic FPV aerial flight over the African savanna at golden hour, the camera flying low and fast above tall yellow grasses dotted with acacia trees. Herds of elephants and giraffes move slowly across the plains\ldots \\
\midrule
A dynamic FPV aerial flight through a vibrant underwater coral city, where colorful corals line the streets and ancient stone ruins rise from the seabed. The camera moves swiftly between schools of tropical fish\ldots \\
\midrule
A sweeping aerial drone shot above the Scottish Highlands, the camera gliding over rolling green moors, mirror-still lochs, and the ruins of an ancient stone castle perched on a small island in the water\ldots \\
\bottomrule
\end{tabular}
\end{table}

\section{Supplementary Video Results}
\label{app:videos}

We provide the following video results in the project webpage at \url{adastate.github.io}, complementing the figures in the main paper.

\paragraph{5-second generation.}
AdaState generations within the training horizon (21 frames), spanning diverse prompts including close-up nature shots, character animation, fantasy scenes, and stylized content. These demonstrate that AdaState produces coherent scene progression with natural camera motion and subject action at the within-horizon regime.

\paragraph{12-second generation.}
AdaState generations beyond the training horizon (51 frames), including cinematic portraits, aerial views, and detailed close-ups. These demonstrate that AdaState maintains temporal coherence and scene progression well past the 5-second training horizon without drift or freezing.

\paragraph{Ablation: state size and window.}
Side-by-side comparisons on two prompts showing three configurations: $F_s{=}3$ with $W_p{+}F{=}9$ (more state slots, wider window), $F_s{=}1$ with $W_p{+}F{=}9$ (single slot, wider window), and $F_s{=}1$ with $W_p{+}F{=}6$ (single slot, tight window, our final configuration). The videos confirm the proportional design rule from Table~2: concentrating the state at a single recurrent position in a tight window produces the strongest dynamics.

\paragraph{Ablation: horizon weight $\alpha$.}
Two panels contrasting short-horizon and long-horizon behavior. At 5 seconds, $\alpha{=}2$ produces the richest dynamics (matching its peak in Table~2). At 30 seconds, $\alpha{=}4$ yields the most stable long-horizon quality, confirming the two-regime training strategy described in Section~4.3.

\paragraph{30-second comparisons.}
Side-by-side comparison panels at 6$\times$ the training horizon (120 frames), each showing four methods spanning the anchor categories: no anchor, static anchor, EMA anchor, and AdaState. The prompts are aerial drone shots over coastal cliffs, an underwater coral city, and a cyberpunk megacity at night. These comparisons make the consistency-dynamics tradeoff from Figure~5 visually concrete: baselines either freeze the scene (static/EMA anchors) or accumulate artifacts (no anchor), while AdaState alone maintains both dynamics and coherence across the full 30-second rollout.

\section{Attention Analysis Details}
\label{app:attn}

This section provides the full methodology for the attention mass analysis summarized in Section~3.1 and visualized in Figure~2(a) of the main paper.

\paragraph{Notation.}
Let $\mathcal{A}^{\ell,h}_{q,k}$ denote the post-softmax attention weight from query token $q$ to key token $k$ at layer $\ell$ and head $h$. Let $L_{\mathrm{frm}}$ denote the number of tokens per frame (1560 in our setting). At generation chunk $c$, the visible K-window covers $F_c$ frames; the $f$-th K-frame spans tokens $k \in [f \cdot L_{\mathrm{frm}},\; (f{+}1) \cdot L_{\mathrm{frm}})$.

\paragraph{Per-record frame mass.}
Each record corresponds to a unique $(\ell, c, t, p)$ tuple of layer, chunk, denoising step, and prompt. For a given record, the attention mass on K-frame $f$, averaged over heads $H$ and a uniformly-spaced query subsample $\mathcal{Q}$ of size $|\mathcal{Q}|{=}32$, is:
\begin{equation}
    \bar{m}^{\ell,c,t,p}_{f} = \frac{1}{|\mathcal{Q}| \cdot H} \sum_{q \in \mathcal{Q}} \sum_{h=1}^{H} \sum_{k=f \cdot L_{\mathrm{frm}}}^{(f+1) \cdot L_{\mathrm{frm}} - 1} \mathcal{A}^{\ell,h}_{q,k}
\end{equation}
By construction, $\sum_{f=0}^{F_c - 1} \bar{m}^{\ell,c,t,p}_{f} = 1$.

\paragraph{Off-diagonal partition.}
The current chunk occupies the last $n_{\mathrm{self}}{=}3$ K-frames. The off-diagonal (cached past) covers the first $n^{(c)}_{\mathrm{off}} = F_c - n_{\mathrm{self}}$ K-frames, i.e.\ the cache depth. We define the off-diagonal mass as $M^{\ell,c,t,p}_{\mathrm{off}} = \sum_{f=0}^{n^{(c)}_{\mathrm{off}}-1} \bar{m}^{\ell,c,t,p}_{f}$ and compute the renormalized off-diagonal share:
\begin{equation}
    s^{\ell,c,t,p}_{f} = \frac{\bar{m}^{\ell,c,t,p}_{f}}{M^{\ell,c,t,p}_{\mathrm{off}}}, \qquad \sum_{f=0}^{n^{(c)}_{\mathrm{off}}-1} s^{\ell,c,t,p}_{f} = 1
\end{equation}
The renormalization treats the entire current chunk as self-attention (rather than excluding only the diagonal element), placing all cached positions on equal footing regardless of how much total mass goes to self-attention.

\paragraph{Aggregation.}
For cache depth $n^* \in \{3, 6, 9, 12, 15, 18\}$, we aggregate over the record set $\mathcal{R}_{n^*} = \{(\ell, c, t, p) : n^{(c)}_{\mathrm{off}} = n^*\}$:
\begin{equation}
    \bar{s}_{n^*,f} = \frac{1}{|\mathcal{R}_{n^*}|} \sum_{(\ell, c, t, p) \in \mathcal{R}_{n^*}} s^{\ell,c,t,p}_{f}, \qquad f \in [0, n^*)
\end{equation}
Records at $t{=}0$ (the cache rerun with clean inputs) are excluded, as they reflect a different attention regime from sampling-time noisy queries.

\paragraph{Analysis Setting.}
We probe the Self-Forcing DMD checkpoint with $P{=}5$ prompts, 21-frame generation (7 chunks), chunk size $n_{\mathrm{self}}{=}3$, $K{=}4$ denoising steps, $L{=}30$ layers, $H{=}12$ heads averaged, and query subsample $|\mathcal{Q}|{=}32$. Total records with cache-rerun excluded: 3{,}600 (600 per cache depth). The anchor rank is consistent across layers (mean rank 2--3, std $< 1$), confirming the pattern is not driven by a subset of layers.

\section{Limitations and Future Work}
\label{app:limitations}

We briefly discuss the limitations in the main paper. Here we expand on two directions.

\paragraph{State capacity.}
Our ablations (Table~2) show that $F_s{=}1$ is sufficient at the current training horizon of 21 frames. However, as rollout length grows, a single frame may not carry enough information to serve as a complete scene reference, particularly for scenes with multiple actors, concurrent events, or parallel narrative threads. Future work could explore multi-slot states with structured roles, or augment the adaptive state with explicit external memory that the model can read and write through attention.

\paragraph{Broader applicability.}
The attention-sink pattern that motivates AdaState has been characterized broadly in causal transformers~\cite{xiao2024efficient}, suggesting the adaptive state formulation could apply to other autoregressive generation domains (e.g.\ audio, 3D, or long-context language generation) wherever a fixed anchor position dominates attention. We leave cross-domain investigation as future work.

\section{Broader Impact and Ethics}
\label{app:ethics}
 
\paragraph{Broader impact.}
AdaState advances streaming video generation by enabling richer temporal dynamics, camera motion, scene progression, and subject action, within a lightweight fine-tuning framework. This has direct applications in creative tools, real-time content generation, and interactive media, where coherent long-horizon video is currently limited by the consistency-dynamics tradeoff. The method requires only 1000 fine-tuning iterations on two GPUs, making it accessible to academic and small-scale research groups.
 
\paragraph{Potential risks.}
As with any improvement in generative video quality, increased realism and temporal coherence could lower the barrier for producing disinformation or non-consensual deepfake content. We emphasize that AdaState does not introduce new generation capabilities, it improves the dynamics of an existing open-source model (Wan 2.1-1.3B), and the same safeguards applicable to the base model (content filtering, watermarking, usage policies) remain applicable. We encourage the community to develop and deploy provenance-tracking and watermarking tools alongside advances in generation quality.
 
\paragraph{Data and human subjects.}
Our training uses only text prompts from publicly available benchmarks (MovieGenBench~\cite{polyak2024movie}) and does not involve any personal, sensitive, or copyrighted visual data. The user study was conducted on the Prolific platform with informed consent; participants were compensated at a rate above the platform's recommended minimum. No personally identifiable information was collected beyond anonymous Prolific IDs.

\paragraph{LLM usage.}
We used large language models to assist with standard engineering tasks such as codebase setup, experiment scripting, and evaluation pipelines. LLMs were not used in the design of the core method, the formulation of the adaptive state mechanism, or the analysis of experimental results.

\end{document}